\documentclass[a4paper,fleqn]{cas-dc}

\usepackage[numbers]{natbib}
\usepackage{algorithm}
\usepackage{algpseudocode}
\usepackage{enumitem}
\newcolumntype{L}{>{\color{black}}l} 

\def\tsc#1{\csdef{#1}{\textsc{\lowercase{#1}}\xspace}}
\tsc{WGM}
\tsc{QE}

\begin{document}
\let\WriteBookmarks\relax
\def\floatpagepagefraction{1}
\def\textpagefraction{.001}

\shorttitle{}    

\shortauthors{Kamel et~al.}  


\title [mode = title]{3D Human Pose Estimation via Spatial Graph Order Attention and Temporal Body Aware Transformer}



%

\author[1]{Kamel Aouaidjia}[orcid=0000-0001-6286-9527]



\ead{kamel@henu.edu.cn}


\credit{Conceptualization, Experiments, Writing original draft}


\affiliation[1]{organization={School of Computer
and Information Engineering, Henan University},
            city={Kaifeng},
            postcode={475001}, 
            state={Henan},
            country={China}}


\author[1]{Aofan Li}


\ead{henulaf@henu.edu.cn}


\credit{Implementation, Experiments}

\author[1]{Wenhao Zhang}


\ead{zxs77889@henu.edu.cn}


\credit{Implementation, Experiments}

\author[1]{Chongsheng Zhang}


\ead{cszhang@ieee.org}
\cormark[1]

\credit{Writing, Reviewing and Editing}

\cortext[1]{Corresponding author.
E-mail: cszhang@ieee.org  (C. Zhang)}



\begin{abstract}
Nowadays, Transformers and Graph Convolutional Networks (GCNs) are the prevailing techniques for 3D human pose estimation. However, Transformer-based methods either ignore the spatial neighborhood relationships between the joints when used for skeleton representations or disregard the local temporal patterns of the local joint movements in skeleton sequence modeling,  while GCN-based methods often neglect the need for pose-specific representations. To address these problems, we propose a new method that exploits the graph modeling capability of GCN to represent each skeleton with multiple graphs of different orders, incorporated with a newly introduced  Graph Order Attention module that dynamically emphasizes the most representative orders for each joint. The resulting spatial features of the sequence are further processed using a  proposed temporal Body Aware Transformer that models the global body feature dependencies in the sequence with awareness of the local inter-skeleton feature dependencies of joints. Given that our 3D pose output aligns with the central 2D pose in the sequence, we improve the self-attention mechanism to be aware of the central pose while diminishing its focus gradually towards the first and the last poses. Extensive experiments on Human3.6m, MPI-INF-3DHP, and HumanEva-I datasets demonstrate the effectiveness of the proposed method. Code and models are made available on Github\footnotemark.

\end{abstract}








\maketitle


\section{Introduction}\label{section1}
Human pose estimation plays a vital role in many applications such as action recognition, robotics, and virtual reality \cite{pang2023skeleton,meng2023motion,kang2023ips}. In the past, obtaining 3D body joint locations required the use of expensive motion capture systems equipped with multiple cameras, wearable sensors, and infrared devices. In recent years, the trend has been to exclusively use images and videos for 3D pose estimation, owing to the advancement of deep learning-based approaches in computer vision. Existing 3D pose estimation approaches fall into two main categories: direct pose regression and 2D to 3D lifting approaches. The first category directly predicts the 3D body joint locations from the input images/videos \cite{li2015maximum,sun2017compositional}, but suffers from the challenge of depth ambiguity. To address this challenge, methods in the second category first leverage the high performance of off-the-shelf 2D pose prediction models to estimate the 2D pose from a single image. Then, they predict the 3D joint positions using a 2D-to-3D pose estimation model \cite{wang2018drpose3d, zhao2023poseformerv2}. In general, 2D to 3D lifting approaches achieve significantly better performance than direct pose regression methods. Moreover, it has been demonstrated that using a sequence of 2D skeletons as input is superior to a single 2D input in predicting the 3D pose.  


Mining spatial-temporal dependencies in a skeleton sequence is crucial for a comprehensive representation of global motion and individual poses. Given that the skeleton structure can be essentially represented as a graph with joints as nodes and bones as edges, the latest techniques utilize Graph Convolutional Networks (GCNs) \cite{kipf2016semi} to effectively model the connections between joints \cite{pavllo20193d, zhao2019semantic}. Unlike MLPs, GCNs explicitly encode kinematic chains (e.g., shoulder, elbow, wrist), to preserve skeletal hierarchy and learns biomechanical constraints such as joints and bones movement limits. Moreover, sparse connectivity fits the skeletal connections, which reduces redundant computations. This facilitates both feature extraction and sharing among joints to create a comprehensive pose representation. 

Alternative methods leverage the self-attention mechanism \cite{vaswani2017attention} inherent in Transformers to selectively focus on crucial joints within each pose \cite{lutz2022jointformer}. Unlike GCNs, which focus on fixed skeletal connections, Transformers use self-attention to dynamically model dependencies across all joints and time steps. This enables them to capture complex motion patterns. To capture both the spatial and temporal dependencies, GCN and Transformer are integrated, in which GCN takes into account the inter-skeleton graph connections between joints throughout the sequence, while the self-attention mechanism assigns different levels of importance to individual elements within the sequence \cite{zheng20213d, li2022mhformer}.

\begin{figure*}
\centering
\includegraphics[width=1\linewidth]{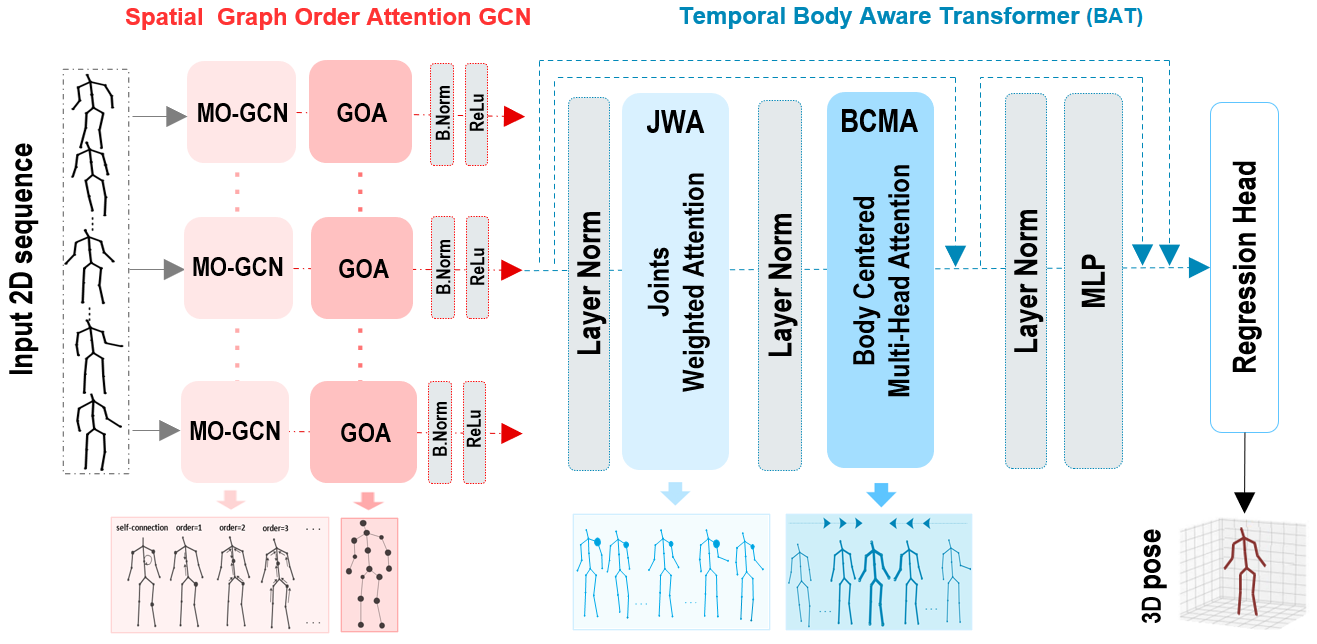}
\caption{Framework of the proposed method. Multiple-order GCN (MO-GCN) generates features of various graph orders for each skeleton. The graph Order Attention (GOA) module assigns weights to various orders for each joint within each pose. The temporal Body Aware Transformer (BAT) captures local attention among body joints using Joints Weighted Attention (JWA) and incorporates global-centered attention using Body-Centred Multi-head Attention (BCMA).}
\label{fig:frame} 
\end{figure*}

However, GCN methods typically rely on a static topology for graph representation. While efforts have been made to integrate multiple skeleton representations using high-order GCNs \cite{zou2020high, wu2022high}, leveraging multiple-order features for joint representation can introduce noise. This is because each joint is better characterized by a specific order that effectively captures its neighboring features within a pose. Additionally, using graph representations to model temporal dependencies poses challenges, as the diverse inter-skeleton joint connections across frames can lead to confusion \cite{wang2020motion, hu2021conditional}. This complexity emerges due to the absence of an inherent connection similar to intra-skeleton connections within the same skeleton. On the other hand, self-attention based methods process the input as a sequence \cite{zheng20213d, li2022mhformer, zhao2023poseformerv2}, ignoring the fact that the joints are not sequentially linked; instead, a joint can be linked to multiple others, where the graph representation is more convenient to represent the spatial connections. Nevertheless, the efficacy of self-attention lies in its power in sequential modeling of the sequential frame features \cite{zhang2022mixste}.

Considering the previous limitations, in this paper, we propose a 3D human pose estimation method that simultaneously leverages the strength of GCN for intra-skeleton spatial relationship modeling and the capability of the attention mechanism for inter-skeleton temporal modeling. Our approach enhances the spatial representation of individual skeleton poses and improves the temporal modeling of pose sequences of existing methods. Following \cite{li2022mhformer, zhao2023poseformerv2}, our input is a sequence of 2D poses, and the output is a single predicted 3D pose aligned with the central frame in the input sequence. To alleviate the problem of using a static topology to represent different poses, we use multiple-order GCNs to model each pose. In contrast to the existing methods that use only the highest order or incorporate all the orders together \cite{zou2020high, Quan2021HigherOrderIF}, we introduce a new learnable module called Graph Order Attention (GOA) that can dynamically focus on the most representative orders for each joint. 

\footnotetext[1]{\href{https://github.com/adjkamel/GOA_BAT}{\url{https://github.com/adjkamel/GOA_BAT}}}

After obtaining the graph embedding features of the skeletons, processing sequential skeleton features directly with the self-attention mechanism, similar to other methods, can model the temporal evolution of the global body features among skeletons, but this ignores the inherent local informative cues of joint dependencies across frames. Moreover, vanilla self-attention methods treat all the skeletons in the sequence equally, without being able to refer to the middle of the sequence \cite{zheng20213d,li2022mhformer, zhao2023poseformerv2}. To address such issues, we propose a new temporal Body Aware Transformer (BAT) that incorporates Joints Weighted Attention (JWA)  on each joint across frames to model its singleton temporal evolution, which is beneficial for local focus on the most informative frames for each joint. Furthermore, we optimize the self-attention mechanism to be aware of the central pose and gradually diminish its attention as it moves away from the central frame, through a new introduced Body-Centred Multi-Head Attention (BCMA). The experimental results and ablation study demonstrate the effectiveness of our method on three public benchmarks: \textit{Human3.6m}, MPI-INF-3DHP, and \textit{HumanEva-I} datasets. The main contributions of this paper can be summarized as follows:

\begin{itemize}
\item We propose an improved 2D-to-3D human pose estimation framework that enhances existing Graph Convolution Networks (GCNs) for better modeling of the spatial relationships between joints and refines transformer attention mechanisms to more effectively capture the temporal evolution of poses across frames.

\item We present a novel Graph Order Attention (GOA) module that determines the optimal graph order for each joint in the skeleton by assigning higher attention weights to the more representative orders.

\item  We propose a new temporal Body Aware Transformer (BAT) that models the temporal dependencies within the sequence with awareness of the feature evolution of each joint across the frames using Joints Weighted Attention (JWA), in the meantime, it emphasizes global body features with more focus on the central frames of the sequence to be aligned with the 3D output using a Body-Centred Multi-Head Attention (BCMA).

\end{itemize}

The rest of this paper is organized as follows: Section II reviews related works. Section III introduces the technical details of our method. Section IV presents the analysis of the experimental results and ablation study, followed by a discussion and conclusion in Section V.

\section{Related work}  \label{section2}

\subsection{GCN in 3D Human Pose Estimation} 
While the skeleton can be represented as a graph, several successful 2D to 3D lifting methods adopted GCN to model the human pose. The first successful attempt to represent the skeleton as a graph was mainly proposed for skeleton-based action recognition \cite{yan2018spatial},  where spatial dependencies are modeled as the intra-body edges between joints in the same frame and the temporal evolution of action is modeled via the inter-frame edges connection between consecutive frames. Later on, GCN was integrated in different ways into the learning process for 3D human pose estimation. \citet{zhao2019semantic} addressed the problems of weight sharing among all nodes and one-step neighborhood feature aggregation. They integrated CNN filters into the graph to acquire distinct weights and capture global and local semantic relationships. \citet{ci2019optimizing} also addressed the problem of weight sharing by introducing a Locally Connected Network (LCN) that utilizes a combination of a fully connected network and a graph network to allow learning a different weight for each node. \citet{choi2020pose2mesh} proposed Pose2Mesh, a  GCN architecture designed to directly estimate the 3D joint locations based on first, a PosNet model lifts 2D to 3D pose, then a second MeshNet takes both 2D and 3D poses to generate 3D human body mesh. \citet{wang2020motion} proposed a spatio-temporal U-shaped GCN that learns motion information alongside 3D joint locations, using a motion loss to incorporate additional supervision from the motion. High-order GCN is proposed by \citet{zou2020high} and \cite{quan2021higher} to model the skeleton by considering multiple orders of connections, allowing for the aggregation of features from distant nodes to achieve more efficient representations. \citet{hu2021conditional} suggest that representing the skeleton as a directed graph is more suitable for capturing its hierarchical structure. They proposed a U-shaped Conditional Directed Graph Convolutional Network that adapts to different graph topologies for various poses. Inspired by Stacked Hourglass Networks for 2D pose estimation \cite{newell2016stacked, xu2021graph} proposed Graph Stacked Hourglass Networks which involve multi-stage refinement within the model to update the 3D joint locations at each stage using encoder-decoder blocks. The encoder down-scales the graph input to a smaller graph size using skeleton pooling layers, while up-pooling layers are used to restore the down-scaled graph to its original size.

\begin{figure*}
\centering
\includegraphics[width=1\linewidth]{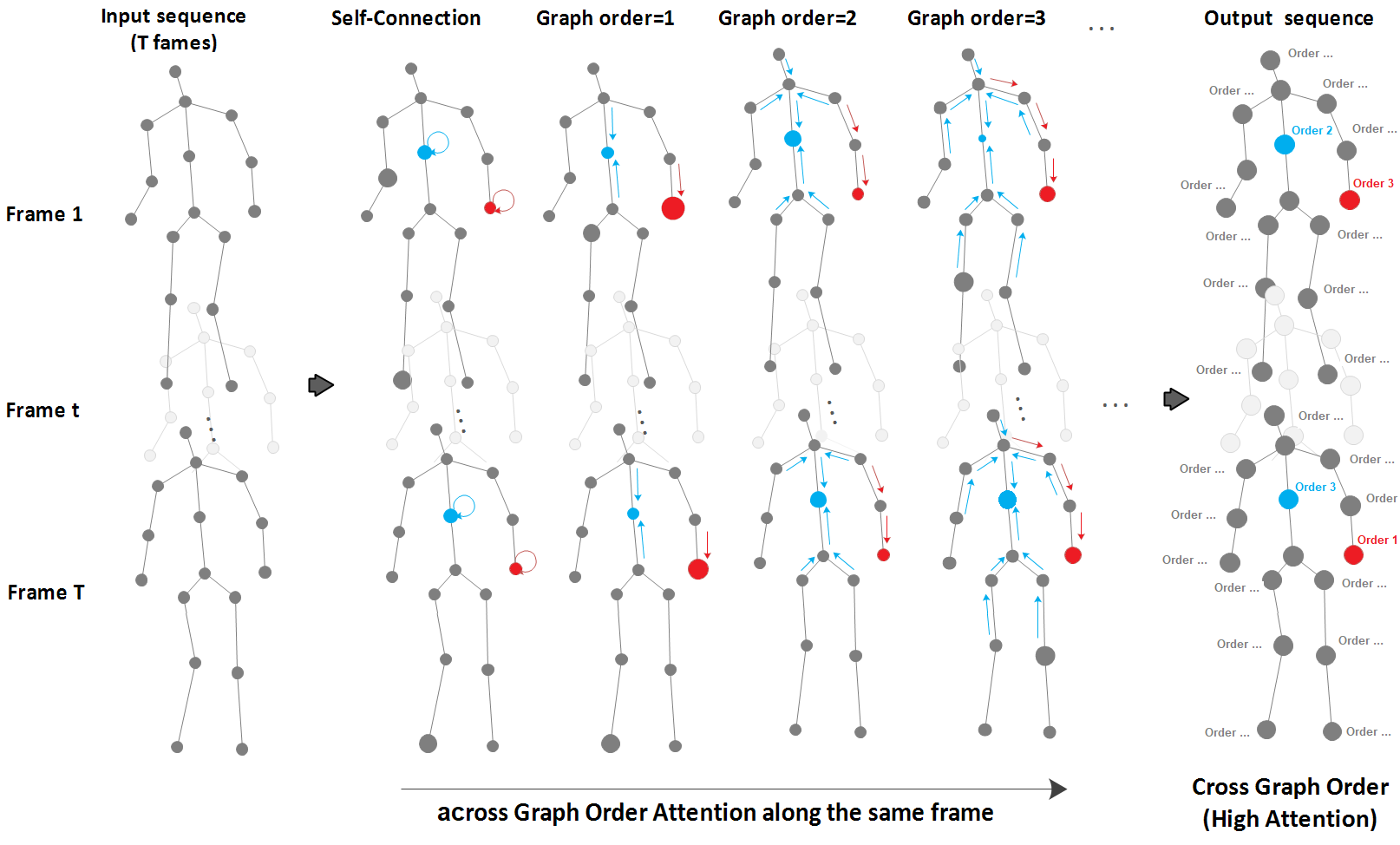}
\caption{Visual illustration of the proposed Graph Order Attention module. For each skeleton in the input sequence, each joint is represented with features of the highest attention weight of the same frame along different orders. Big joints represent high attention weights.}
\label{fig:scog} 
\end{figure*}

\subsection{Transformer in 3D Human Pose Estimation}  
In the task of 2D to 3D pose estimation, the skeleton can be represented as a sequence of joints, and the temporal evolution of the skeleton poses over time can also be represented as a sequence. For both tasks, a Transformer with self-attention was utilized for sequential modeling. \citet{lin2021end} introduced a framework that utilizes a combination of Transformer and CNN, specifically designed for the simultaneous recovery of human mesh and pose estimation from a single image. \citet{zheng20213d} (Posformer) is among the leading works for 3D pose estimation using Transformer. It utilizes a sequence of frames to predict a single 3D pose using two Transformers, a spatial transformer that processes each skeleton input as a sequence of joints to extract spatial features, and then a temporal Transformer processes the resulting spatial features sequence to predict the 3D pose. To reduce the sequence redundancy, computation cost, and aggregate information from local contexts, \citet{li2022exploiting} introduced Strided Transformer Encoder (STE) to gradually reduce the temporal dimension in the vanilla Transformer encoder by replacing the fully connected layers in the feed-forward network block with strided convolutions. The Multi-Hypothesis Transformer (MHFormer), introduced by \citet{li2022mhformer}, is designed to acquire spatio-temporal representations by considering multiple generated hypotheses of the 3D pose, and then cross-hypothesis features are learned from the different representations to generate the final 3D pose. The Jointformer proposed by \citet{lutz2022jointformer} employs two consecutive Transformers, a Joint Transformer estimates an initial 3D pose and generate a prediction error, and then a Refinement Transformer enhances the pose accuracy using both the initial prediction and the prediction error.

\section{Methodology}
The proposed learning architecture is illustrated in Fig. \ref{fig:frame}. Given the input skeleton sequence of 2D poses, a spatial GCN consists of a  Multi-Order GCN (MO-GCN) followed by a Graph Order Attention (GOA) to extract features for each skeleton. Specifically, MO-GCN aggregates features from various distances (orders) of neighboring joints, and a GOA assigns importance to different orders for each joint in every pose. The resulting sequence features are further processed through a temporal Body Aware Transformer (BAT) which integrates two attention mechanisms: a Joints Weighted Attention captures local feature dependencies among joints, followed by a Body-Centred Multi-head Attention (BCMA), which captures global body features, focusing on the central frames. In the following sections, we provide a detailed description of each component of the proposed method.

\subsection{Spatial GCN}

\subsubsection{Multiple-Order GCN (MO-GCN)}
Let the 2D skeleton sequence  $S=\{S_1, ,...,S_T\}$ generated by off-the-shelf 2D detection model, where $S \in \mathbb{R}^{T \times J \times 2}$, $T$ the number of frames, $J$ the number of 2D joints per frame. Following the graph modeling \cite{kipf2016semi}, each skeleton can be considered as a graph $S_i=\{ V,E \}$ where $V$ is the set of joints (nodes) and $E$ are the set of edges. The symmetric adjacency matrix is defined as $\textbf{A} \in \mathbb{R}^{J \times J}$ where $\textbf{A}_{ij}=\{0,1\}$ encodes the edges between neighboring joints, where $1$ means there is an edge between the joints $i$ and $j$ and $0$ for no edge.  A graph of order $k$ considers the neighboring k-hop connections between joints \cite{zou2020high}, as well as the (k-1)-hop connections. The adjacency matrix  $\textbf{A}^{k}_{ij}=\{0,1\}$ encodes the edges between k-hop neighboring joints, where $1$ means there is $k$ edge between the joints $i$ and $j$. We consider a graph of order $k=0$ as the self-connection of a joint with itself where $\textbf{A}^{0}_{ij}=I$ ($I$ the identity matrix). To spatially extract features from an individual skeleton $S_i$, we define a GCN of order $k$ without an activation function as: 
\begin{equation}
  S^{k}_{i} = \tilde{\textbf{A}}^k  \: S_i \: \textbf{W}^{k}
\end{equation}
Where $\tilde{\textbf{A}}$ is the normalized adjacency matrix similar to \citet{kipf2016semi}, $\textbf{W}^{k} \in \mathbb{R}^{J \times D} $ is a learnable weight for the order k features of the skeleton $S_i$. $D$ is the features dimension, and $S^{k}_{i}$ is the resulting graph embedding features of order $k$ of the skeleton $S_i$, where $S^{k}_{i}\in \mathbb{R}^{J \times D}$

\begin{figure*}
\centering
\includegraphics[width=0.7\linewidth]{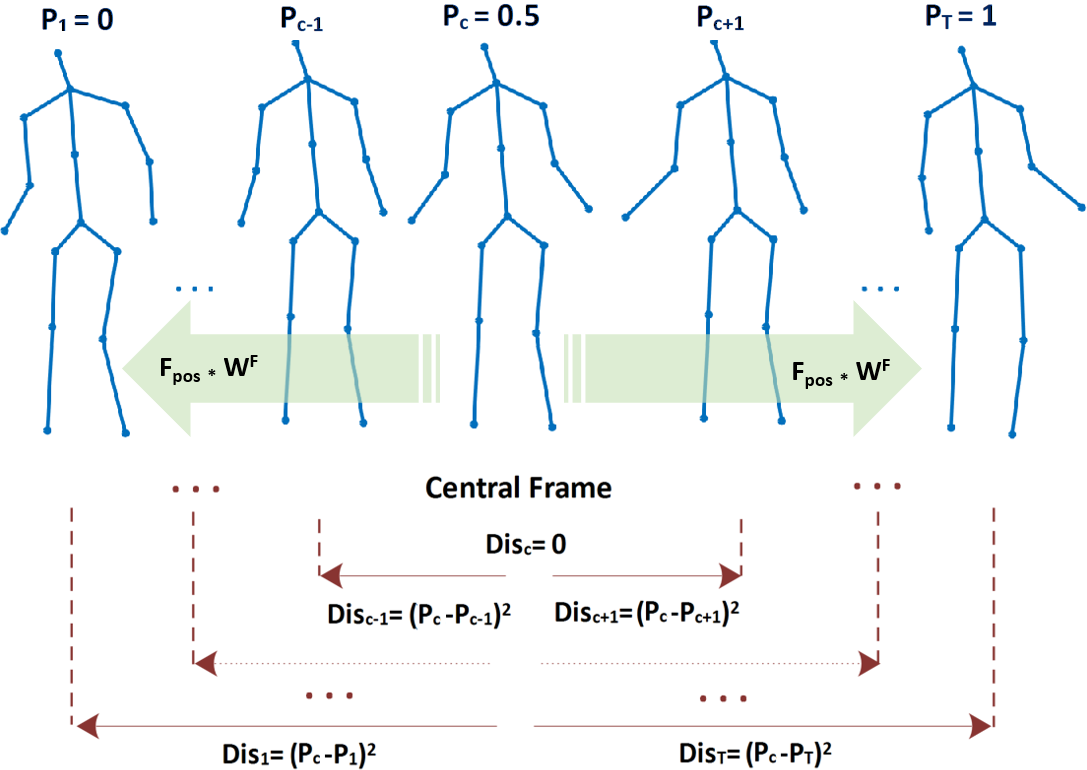}
\caption{Illustration of the distance calculation between the central frame and the left/right poses in the sequence. $Pos = [P_1, \dots, P_c, \dots, P_T]$, $Dis = [Dis_1, \dots, Dis_c, \dots, Dis_T]$, and $F_{pos} * \textbf{W}^{F}$ correspond to those defined in equations \ref{eq:dis} and \ref{eq:Pscl}.}
\label{fig:scog} 
\end{figure*}

\subsubsection{Graph Order Attention (GOA) Module}
Each skeleton $S_i$ of the sequence is represented by different graph order features $S^k_i$, where $k=\{0,1,\ldots , R\}$ and $R$ is the highest order. The trivial way to represent the skeleton using multiple graph orders is to concatenate all the order features for further processing. However, some of the joint embedding features could be more representative in a specific order than the others, which makes using the concatenation leads to inefficient joint representation. To alleviate such confusion, we introduce a learnable order attention mechanism that selectively assigns higher attention weights to specific graph orders for each joint. The proposed attention is based on weighing the joint features along the orders for each skeleton. Fig. \ref{fig:scog} illustrates visually the GOA module. We denote the concatenation of the graph orders features of each skeleton $S_i$ as:
\begin{equation}
  OrdS_{i} = S^0_i \mathbin\Vert S^1_i \mathbin\Vert 	\ldots \mathbin\Vert S^R_i 
\end{equation}
Where $\mathbin\Vert$ is the concatenation along a new dimension determined by the number of orders $R+1$, and $OrdS_{i} \in \mathbb{R}^{J \times (R+1) \times D}$. We set $Q$, $K$ as learnable query and key of our attention mechanism, with the weights $\textbf{W}_{q},\textbf{W}_{k} \in\mathbb{R}^{D\times D}$, respectively, where:
\begin{align}
  Q &= OrdS_{i} \: \textbf{W}_{q} \label{eq:3}  \\     
  K &= OrdS_{i} \: \textbf{W}_{k} \nonumber
\end{align}
We define a learnable  order weighted attention $Ott$ as: 
\begin{equation}
  Ott = softmax(\tanh(Q+K) \: \textbf{W}_{o})
\end{equation}
Where $Ott \in  \mathbb{R}^{J \times (R+1)}$ represents the attention weights for each joint in each order, with $\textbf{W}_{o} \in\mathbb{R}^{D}$. The resulting $Ott$ can also be written as :  
\begin{equation}
Ott = Ott_0 \mathbin\Vert Ott_1 \mathbin\Vert \dots \mathbin\Vert Ott_R
\end{equation}
Where $Ott_k$ are the attention weights for the skeleton joints in order $k$. Finally, the GOA module is calculated by:
\begin{align}
   GOA &= sum(OrdS_{i} \odot  Ott) = \sum_{k=0}^{R}S^k_i  \: Ott_k \\
   \tilde{S}_{i} &= \sigma (GOA)  \nonumber
\end{align}
Where $OrdS_{i}  \odot Ott$ is the element-wise multiplication of the attention weights by the features of each joint across the orders, and $GOA$ is the weighted sum across the orders. $\tilde{S}_{i} \in  \mathbb{R}^{J \times D} $ is the new skeleton features resulting from applying $\sigma$ activation function on the $GOA$.

\begin{figure*}
\centering
\includegraphics[width=1\linewidth]{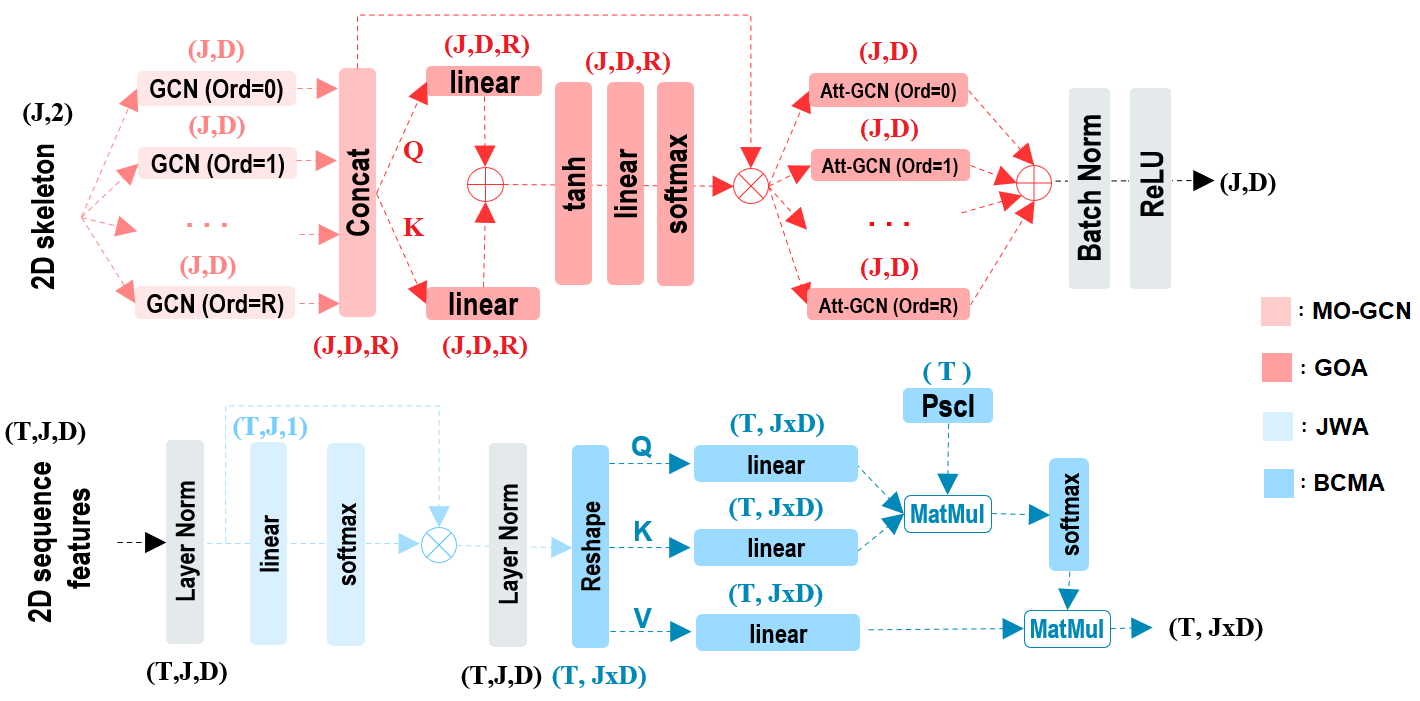}
\caption{Detailed structure of GOA, JWA, and BCMA. T: Number of frames, J: Number of joints, R: Number of the highest order, D: Features dimension. Pscl: Scaling position vector in Equation \ref{eq:Pscl}. The colors match the components in Fig. \ref{fig:frame}. Q, K (red) correspond to the symbols in Equation \ref{eq:3}, and Q, K, V (blue) correspond to the symbols in Equation \ref{eq:13}. }
\label{fig:modules} 
\end{figure*}
\subsection{Temporal Body Aware Transformer (BAT)}
For sequential modeling, a Transformer equipped with the self-attention mechanism has demonstrated exceptional predictive performance. The encoder of the vanilla Transformer involves applying feature embedding to the input, followed by positional embedding, self-attention, and a Multi-Layer Perceptron (MLP). However, applying such a structure to the input features of the sequence tends to focus only on global body features, neglecting attention to local features such as joints. Moreover,  it applies attention to all the elements in the sequence equally, while in our case, the central frames align with the 3D pose more than the rest of the elements of the sequence. Therefore, we introduce a new Transformer structure equipped with joint weighted attention for each joint across frames, followed by a self-attention mechanism reinforced with a central focus on the input sequence features.

\subsubsection{Joints Weighted Attention (JWA)}
Our Joints Weighted Attention (JWA) processes individual joints' temporal evolution independently through frame-wise attention. By applying separate attention mechanisms to each joint, the model can precisely capture unique motion patterns without interference from unrelated body parts. This local attention creates dedicated temporal filters for each joint, rather than treating all joints as equally relevant and introducing noise. Additionally, our JWA is considered complementary to global attention, as it focuses on individual joints' evolution while global attention captures overall body motion.

The application of MO-GCN followed by the GOA module to each skeleton generates a new sequence of graph embedding features $\tilde{S}_1,\dots,\tilde{S}_T$. We consider the concatenation of the graph embedding feature sequence as follows:
\begin{equation}
  \tilde{S} = \tilde{S}_0 \mathbin\Vert \tilde{S}_1 \mathbin\Vert 	\ldots \mathbin\Vert \tilde{S}_T 
\end{equation}
Where $\mathbin\Vert$ is the concatenation along new dimension determined by the number of frames $T$, and hence, $\tilde{S} \in \mathbb{R}^{T \times J \times D} 
$. We define the attention weights of the joints across the frames  as:
\begin{equation}
  Jtt = softmax(\tilde{S} \: \textbf{W}^{J})
\end{equation}
With $ Jtt, \textbf{W}^{J} \in \mathbb{R}^{T \times J}$. Where  $Jtt$ can be written as:
\begin{equation}
Jtt = Jtt_0 \mathbin\Vert Jtt_1 \mathbin\Vert \dots \mathbin\Vert Jtt_T
\end{equation}
The new skeleton sequence features $\tilde{S}^{J} \in \mathbb{R}^{T \times J \times D}$  which is the result of the JWA is calculated as:
\begin{align}
  \tilde{S}^{J} &=  \tilde{S} \odot Jtt \\
  &=    (\tilde{S}_0  \mathbin\Vert 	\ldots \mathbin\Vert \tilde{S}_T) \odot (Jtt_0  \mathbin\Vert \dots \mathbin\Vert Jtt_T) \nonumber
\end{align}
Where $\tilde{S} \odot Jtt$ is the element-wise multiplication of the attention weights by the features joints across the frames.

\subsubsection{Body-Centred Multi-Head Attention (BCMA)}
We further apply a multi-head self-attention mechanism to capture dependencies between global body features in the sequence. The skeleton sequence features $\tilde{S}^{J} \in \mathbb{R}^{T \times J \times D}$ can be expressed as  $\tilde{S}^{J} \in \mathbb{R}^{T \times G}$, where $G=J \times D$ encapsulates the joint embedding features from all joints to represents the global features dimension of each skeleton. The vanilla self-attention mechanism \cite{vaswani2017attention}  applied to the input sequence features is defined as:
\begin{equation}
  Att (Q,K,V) = softmax( \frac{Q\:K^T}{\sqrt G})V
\end{equation}
We introduce a learnable central scaling mechanism that focuses on the center of the sequence, diminishing its focus gradually toward the start and the end of the sequence. Initially, we set $Q=K=V=\tilde{S}^{J}$, and we define $Pos=[0,1] \in \mathbb{R}^T$ as the positions of frames in the sequence, where the initial position value is 0 and the last position is 1. We measure the distance $Dis$ of each position from the central position as:
\begin{equation}
Dis = (Pos - P_{c})^2
  \label{eq:dis} 
\end{equation}

Where $P_{c}$ is a positive scalar value representing the center position around which the attention scaling occurs. By default, $P_{c}=0.5$ represents the central frame position. To specify the decay rate of the attention from the center towards the start and end of the sequence, a learnable scale factor $F_{pos}$ with the weight $W^{F}$ is defined. The final scaling position vector $P_{scal}$ is expressed with the following equation:
\begin{equation}
  P_{scl} = exp(-F_{pos} * \textbf{W}^{F} * Dis)
  \label{eq:Pscl} 
\end{equation}
Where the vector $P_{scl} \in \mathbb{R}^T$ is characterized by higher values for positions closer to the center and decreasing values as positions move away from the center. The formulation of the new centered self-attention mechanism is defined as:
\begin{equation}
  CentAtt (Q,K,V) = softmax( \frac{Q\:K^T\:P_{scl}}{\sqrt G})V
    \label{eq:13} 
\end{equation}
As a result, the Body-Centered Multi-head Attention (BCMA) is expressed as:
\begin{align}
 CentMultiHead(Q,K,V) &= Concat(h_1, ..., h_n)\textbf{W}^O
\end{align}
Where $n$ is the number of heads, $\textbf{W}^O$ is a learnable weight of the centered multi-head attention, and $h_i$ is a single centered self-attention head, where:
\begin{align}
 h_i &= CentAtt (Q\textbf{W}_i^Q,K\textbf{W}_i^K,V\textbf{W}_i^V) 
\end{align}

\subsection{Model Architecture}
Fig. \ref{fig:frame} shows the global model structure and Fig. \ref{fig:modules} shows the corresponding main model components presented with the same colors as Fig. \ref{fig:frame}. Each skeleton is treated as a graph, and multiple GCNs of different orders are applied to the same input skeleton (MO-GCN). The features from different GCNs are concatenated and used as input to the GOA module. Two linear layers are created from the concatenation, serving as the query and key of the attention. The summation of these layers is followed by a sequence of operations, including tanh activation, a linear layer, and a softmax, to generate the attention weights which are utilized to selectively weigh the concatenated features from GCNs, emphasizing different orders. The resulting features are split into multiple parts to be aligned with the original sizes of the GCNs' features. The new multiple attention features are summed and followed by batch normalization and ReLu activation to form the output of the GOA module, as illustrated in Fig. \ref{fig:modules}(top).

The previous process is applied to each skeleton in the sequence, yielding a sequence of graph embedding features. The sequence is then fed into the temporal Body Aware Transformer equipped with the two consecutive attention mechanisms JWA and BCMA. Each attention is preceded by a layer normalization, as illustrated in Fig \ref{fig:modules}(bottom). A residual connection links the input sequence and the output of the BCMA (Fig \ref{fig:frame}) followed by a batch normalization and a Multilayer Perceptron (MLP). The MLP is composed of two linear layers with ReLU activation in between.  Another residual connection links the output of BCMA and the output of the MLP. At the end, a regression head consisting of a layer normalization followed by a linear layer is used for the prediction of the 3D joints.

\section{Experiments}

\subsection{Datasets and Evaluation Protocols}
\subsubsection{Human3.6m \cite{ionescu2013human3}} Human3.6m \cite{ionescu2013human3} is a widely used dataset for indoor single-person human pose estimation with 3.6 million 3D human poses and corresponding images. The poses are captured for 11 actors from 4 views performing 17 actions such as Discussion, Taking Photos, Talking Phone, etc., using 15 motion capture devices and body markers. Following the state-of-the-art works \cite{fang2018learning}, \cite{pavllo20193d, zhao2019semantic, liu2020attention, wang2020motion}, we also adopt the same settings for training and evaluation. The subjects S1, S5, S6, S7, S8 are used for training and S9, S11  are used for testing. The evaluation $Protocol\#1$ is the MPJPE (Mean Per Joint Position Error) which refers to the average Euclidean distance between the predicted joint locations and the ground truth, and $Protocol\#2$ is the P-MPJPE which is similar to MPJPE, but it is calculated after aligning the ground truth with the predicted 3D pose using translation, rotation, and scale. Like the other methods, in the conducted experiments on Human3.6m, we utilized the 2D detections produced by CPN (Cascaded Pyramid Network) \cite{chen2018cascaded}, and the ground truth (GT) 2D poses.

\subsubsection{MPI-INF-3DHP \cite{mehta2017monocular}}
MPI-INF-3DHP \cite{mehta2017monocular} is a more challenging dataset including both indoor and outdoor scenes. Its training set consists of 8 subjects performing 8 different activities, while the test set includes 7 activities across three distinct scenes: green screen, non-green screen, and outdoor. Following the protocols in \cite {shan2022p, zheng20213d, chen2021anatomy}, we train our proposed method on all activities captured from 8 camera angles in the training set and evaluate it on valid frames within the test set. Results are reported using the MPJPE.

\subsubsection{HumanEva-I \cite{sigal2010humaneva}}
HumanEva-I \cite{sigal2010humaneva} is a much smaller dataset comprising 7 calibrated video sequences synchronized with 3D body poses obtained from a motion capture system. The database includes recordings of 4 subjects engaged in 6 common actions, such as walking, jogging, gesturing, and more. Following \cite{martinez2017simple, liu2020attention, zheng20213d}, the subjects S1, S2, and S3 with the actions Walking and Jogging are used for training and testing, and $Protocol\#2$ is used for evaluation.

\begin{table*}
\renewcommand{\arraystretch}{1.4}
\setlength{\tabcolsep}{0.6pt}
\small
\caption{Comparison of error  between the predicted 3D pose and the ground truth with the state-of-the-art on Human3.6m dataset using 2D detection (CPN) joints (top) and 2D ground truth joints (bottom) as input under $Protocol\#1$ (MPJPE). † indicates using temporal information.}
\centering
\begin{tabular}{l  c c c c c c c c c c c c c c c | c}
\hline
\bfseries   Method (CPN)    & Dir. & Disc. & Eat. & Greet. & Phone. & Photo. & Pose. & Pur. & Sit. & SitD. & Smoke.&  Wait.&  WalkD.&  Walk. & WalkT.& Avg. \\
\hline

\hline

\citet{martinez2017simple} (ICCV’17)&51.8& 56.2& 58.1& 59.0 &69.5& 78.4& 55.2& 58.1& 74.0& 94.6 &62.3 &59.1&65.1 &49.5 &52.4& 62.9	\\

\citet{fang2018learning} (AAAI’18)&50.1 & 54.3& 57.0 &57.1& 66.6 &73.3& 53.4 &55.7 &72.8 &88.6 &60.3& 57.7& 62.7 &47.5& 50.6 &60.4\\

\citet{pavlakos2018ordinal} (CVPR’18)&48.5& 54.4& 54.4& 52.0& 59.4& 65.3& 49.9 &52.9& 65.8 &71.1& 56.6 &52.9 &60.9 &44.7 &47.8& 56.2 \\

\citet{lee2018propagating} (ECCV’18) &40.2& 49.2& 47.8& 52.6 &50.1 &75.0& 50.2& 43.0 &55.8 &73.9 &54.1 &55.6& 58.2 &43.3& 43.3 &52.8 \\

\citet{zhao2019semantic} (CVPR’19) &47.3 &60.7& 51.4 &60.5& 61.1& 49.9 &47.3 &68.1& 86.2 &\textbf{55.0} &67.8 &61.0& \textbf{42.1}& 60.6 &45.3 &57.6 \\

\citet{ci2019optimizing} (ICCV’19) &46.8 &52.3 &44.7 &50.4& 52.9& 68.9 &49.6& 46.4& 60.2 &78.9& 51.2& 50.0 &54.8& 40.4 &43.3& 52.7 \\

\citet{cai2019exploiting} (ICCV’19) † &44.6& 47.4& 45.6& 48.8& 50.8 &59.0 &47.2 &43.9 &57.9 &61.9 &49.7 &46.6 &51.3 &37.1 &39.4 &48.8\\
\citet{pavllo20193d} (CVPR’19) †  &45.2& 46.7 &43.3 &45.6 &48.1& 55.1& 44.6& 44.3& 57.3& 65.8 &47.1& 44.0& 49.0& 32.8 &33.9 &46.8 \\

\citet{xu2020deep} (CVPR’20) † & \textbf{37.4} &43.5& 42.7 &42.7& 46.6& 59.7& 41.3& 45.1& \underline{52.7}& 60.2 &45.8& 43.1& 47.7& 33.7 &37.1 &45.6 \\
\citet{liu2020attention} (CVPR’20) † & 41.8 & 44.8 & 41.1 & 44.9 & 47.4&  54.1&  43.4&  42.2 & 56.2 & 63.6 & 45.3 & 43.5&  45.3&  31.3 & 32.2 & 45.1\\

\citet{zeng2020srnet} (CVPR’20) † & 46.6 & 47.1 & 43.9 & 41.6 & 45.8 & \underline{49.6} & 46.5 & \underline{40.0} & 53.4 & 61.1 & 46.1 & 42.6 & 43.1 & 31.5 & 32.6 & 44.8\\

\citet{xu2021graph} (CVPR’21) & 45.2 &49.9 &47.5 &50.9& 54.9& 66.1 &48.5 &46.3& 59.7& 71.5& 51.4& 48.6& 53.9& 39.9& 44.1& 51.9 \\
\citet{zhou2021hemlets}  (PAMI’21) &  \underline{38.5} & 45.8 & 40.3 & 54.9 & \textbf{39.5} & \textbf{45.9} & \textbf{39.2}&  43.1 & \textbf{49.2}&  71.1 & \textbf{41.0} & 53.6 & 44.5&  33.2&  34.1 & 45.1 \\
\citet{liu2021graph} (ICRA’21) † & 43.3 & 46.1 & 40.9 & 44.6 & 46.6 & 54.0 & 44.1 & 42.9 & 55.3 & \underline{57.9} & 45.8 & 43.4 & 47.3 & 30.4 & 30.3 & 44.9 \\

\citet{li2022mhformer} (CVPR’22) † & 39.2 & \underline{43.1} & \underline{40.1} & \textbf{40.9} & 44.9 & 51.2 & \underline{40.6} & 41.3&  53.5 & 60.3 & \underline{43.7} & \textbf{41.1}&  43.8 & 29.8 & 30.6 & \underline{43.0}\\
 \citet{shan2022p} (ECCV’22) † &38.9 & \textbf{42.7}  &40.4 &\underline{41.1}  &45.6 & 49.7 & 40.9  &\textbf{39.9 }& 55.5 & 59.4  &44.9 & 42.2  &\underline{42.7}  &\underline{29.4} & \textbf{29.4}  &\textbf{42.8}\\
\citet{wang2022motion} (TCDS’22) † & 42.4 &43.5 &41.0 &43.5& 46.7& 54.6 &42.5 &42.1& 54.9& 60.5& 45.7& 42.1& 46.5& 31.7& 33.7& 44.8 \\
\citet{yu2023gla} (ICCV’23) † &41.3  &44.  &40.8  &41.8  &45.9  &54.1  &42.1  &41.5  &57.8  &62.9  &45.0 & 42.8  &45.9  &\textbf{29.4}  &\underline{29.9} & 44.4\\

Ours (T=324, 6xBAT) † &41.3  & 43.8       & \textbf{39.4} & 41.6    &\underline{44.5}  & 52.0 &42.2           &41.0         & 54.2        &62.8          & 44.0& \underline{42.0}         &44.1 & 30.2 &	30.8 &43.5\\

\hline

\bfseries   Method (GT)    & Dir. & Disc. & Eat. & Greet. & Phone. & Photo. & Pose. & Pur. & Sit. & SitD. & Smoke.&  Wait.&  WalkD.&  Walk. & WalkT.& Avg. \\
\hline

\citet{martinez2017simple} (ICCV’17) & 37.7 & 44.4 & 40.3 & 42.1 & 48.2& 54.9 &44.4 &42.1 &54.6 & 58.0& 45.1 &46.4 &47.6 &36.4& 40.4 &45.5 \\

\citet{lee2018propagating} (ECCV’18)  & 32.1 &36.6 &34.3 &37.8 & 44.5& 49.9& 40.9 &36.2 &44.1 &45.6 &35.3 &35.9 &30.3& 37.6& 35.5 &38.4 \\

\citet{zhao2019semantic} (CVPR’19) & 37.8& 49.4 &37.6& 40.9& 45.1& 41.4 &40.1 &48.3& 50.1 &42.2& 53.5 &44.3& 40.5& 47.3 &39.0& 43.8 \\

\citet{ci2019optimizing} (ICCV’19) & 36.3 & 38.8 & 29.7 & 37.8& 34.6 &42.5& 39.8 &32.5 &36.2 &39.5& 34.4 &38.4 &38.2& 31.3 &34.2& 36.3 \\

\citet{liu2020attention} (CVPR’20) † & 34.5 &37.1 &33.6& 34.2& 32.9& 37.1& 39.6& 35.8 &40.7& 41.4 &33.0& 33.8& 33.0& 26.6 &26.9&34.7 \\

\citet{xu2021graph} (CVPR’21) & 35.8 &38.1 &31.0 &35.3 &35.8 &43.2 &37.3& 31.7 &38.4 &45.5 &35.4 &36.7 &36.8 &27.9 &30.7 &35.8 \\

\citet{zheng20213d} (ICCV’21) † &30.0 &33.6& 29.9 &31.0 &30.2 &33.3 &34.8 &31.4 &37.8 &38.6 &31.7& 31.5 &29.0& 23.3 &23.1& 31.3 \\

\citet{li2022mhformer} (CVPR’22) † & 27.7 & 32.1 & 29.1 & 28.9 & 30.0 & 33.9 & 33.0 & 31.2 & 37.0 & 39.3 & 30.0 & 31.0 & 29.4 & 22.2 & 23.0 & 30.5 \\

\citet{shan2022p} (ECCV’22) † & 28.5 & 30.1 & \underline{28.6} & 27.9 & 29.8 & 33.2 & 31.3 & 27.8 & \underline{36.0} & \underline{37.4} & \underline{29.7} & 29.5 & 28.1 & \underline{21.0} & \underline{21.0} & 29.3 \\

\citet{yu2023gla} (ICCV’23) † & \underline{26.5} & \textbf{27.2} & 29.2 & \underline{25.4} & \underline{28.2} & \underline{31.7} & \underline{29.5} & \underline{26.9} & 37.8 & 39.9 & 29.9 & \textbf{27.0} & \underline{27.3} & \textbf{20.5} & \textbf{20.8} & \underline{28.5} \\

Ours (T=324, 6xBAT) † & \textbf{26.3} & \underline{28.3} & \textbf{25.6} & \textbf{25.3} & \textbf{26.6} & \textbf{29.4} & \textbf{29.1} & \textbf{25.6} & \textbf{31.4} & \textbf{33.8} & \textbf{27.6} & \underline{27.1} & \textbf{26.5} & 21.5 & 21.3 & \textbf{27.0} \\

\hline

\end{tabular}
\label{table:h36p1}
\end{table*}

\begin{table*}
\renewcommand{\arraystretch}{1.4}
\setlength{\tabcolsep}{0.7pt}
\small
\caption{Comparison of error between the predicted 3D pose and the ground truth with the state-of-the-art on Human3.6m dataset using 2D detection (CPN) joints (top) and 2D ground truth joints (bottom) as input under $Protocol\#2$ (MPJPE). † indicates using temporal information.}

\centering
\begin{tabular}{l | c c c c c c c c c c c c c c c | c}

\hline

\bfseries   Method (CPN)   & Dir. & Disc. & Eat. & Greet. & Phone. & Photo. & Pose. & Pur. & Sit. & SitD. & Smoke.&  Wait.&  WalkD.&  Walk. & WalkT.& Avg. \\
\hline
\citet{martinez2017simple} (ICCV'17) & 39.5& 43.2& 46.4& 47.0& 51.0 &56.0 &41.4& 40.6& 56.5& 69.4& 49.2& 45.0& 49.5& 38.0& 43.1 &47.7 \\

\citet{fang2018learning} (AAAI'18) &38.2& 41.7& 43.7& 44.9& 48.5 &55.3& 40.2& 38.2 &54.5& 64.4 &47.2& 44.3& 47.3& 36.7 &41.7 &45.7 \\

\citet{pavlakos2018ordinal} (CVPR'18) &34.7& 39.8& 41.8 &38.6& 42.5& 47.5 &38.0 &36.6& 50.7 &56.8 &42.6 &39.6& 43.9 &32.1& 36.5& 41.8 \\

\citet{lee2018propagating} (ECCV'18)  & 34.9 &35.2 &43.2 &42.6& 46.2 &55.0& 37.6 &38.8 &50.9 &67.3& 48.9& 35.2 &\textbf{31.0} &50.7 &34.6& 43.4 \\

\citet{ci2019optimizing} (ICCV'19) & 35.7 &37.8& 36.9& 40.7& 39.6& 45.2 &37.4& 34.5& 46.9& 50.1& 40.5& 36.1& 41.0 &29.6& 33.2 &39.0 \\

\citet{pavllo20193d} (CVPR'19) †&  34.1 &36.1 &34.4 &37.2 &36.4 &42.2 &34.4 &33.6 &45.0& 52.5 &37.4& 33.8 &37.8 &25.6& 27.3& 36.5 \\

\citet{xu2020deep} (CVPR'20) † & \textbf{31.0}& \textbf{34.8} &34.7 &34.4& 36.2& 43.9 &\underline{31.6}& 33.5& \textbf{42.3}& 49.0& 37.1& 33.0& 39.1& 26.9& 31.9 &36.2 \\
\citet{chen2018cascaded} (ICCV'20) † &32.9 &35.2 &35.6& 34.4& 36.4 &42.7 &\textbf{31.2} &32.5& 45.6 &50.2 &37.3& 32.8 &36.3& 26.0 &\underline{23.9}& 35.5\\

\citet{liu2020attention} (CVPR'20) † & 32.3& 35.2 &33.3& 35.8 &35.9& 41.5& 33.2 &32.7 &44.6 &50.9 &37.0 &32.4 &37.0 &25.2 &27.2& 35.6 \\
 
\citet{shan2021improving} (MM'21) † &32.5& 36.2 &33.2 &35.3 &35.6& 42.1 &32.6 &31.9& \underline{42.6}& \textbf{47.9}& 36.6& \textbf{32.1} &34.8 &24.2 &25.8& 35.0 \\

\citet{shan2022p}  (ECCV'22) † & \underline{31.3} &\underline{35.2}& 32.9 &\textbf{33.9}& 35.4& \textbf{39.3}& 32.5 &\textbf{31.5}& 44.6& \underline{48.2}& 36.3 &32.9 &\underline{34.4} &23.8 &\textbf{23.9} &\textbf{34.4}\\

\citet{yu2023gla} (ICCV'23) † &32.4& 35.3& \textbf{32.6} &\underline{34.2} &\underline{35.0}& 42.1& 32.1& \underline{31.9} &45.5 &49.5 &\underline{36.1}& \underline{32.4}& 35.6 &\textbf{23.5} &24.7& 34.8\\

Ours (T=324, 6xBAT) † &32.0   & 35.6   & \underline{32.8} & 34.4  & \textbf{34.7} & \underline{40.0} & 32.9    & 32.2&  44.2 & 50.5  & \textbf{35.5}& 32.7  &    35.1&  \underline{23.7} & 24.6 &\underline{34.6}\\

\hline
\bfseries   Method (GT)  & Dir. & Disc. & Eat. & Greet. & Phone. & Photo. & Pose. & Pur. & Sit. & SitD. & Smoke.&  Wait.&  WalkD.&  Walk. & WalkT.& Avg. \\
\hline

\citet{ci2019optimizing} (ICCV'19) &24.6 &28.6 &24.0& 27.9 &27.1 &31.0 &28.0& 25.0& 31.2 &35.1 &27.6 &28.0 &29.1 &24.3& 26.9& 27.9 \\

\citet{yu2023gla} (ICCV'23) † &\underline{20.2} &\textbf{21.9} &\underline{21.7} &\textbf{19.9} &\underline{21.6} &\underline{24.7} &\underline{22.5} &\underline{20.8} &\underline{28.6} &\underline{33.1} &\underline{22.7} &\textbf{20.6} &\textbf{20.3}& \underline{15.9}& \underline{16.2} &\underline{22.0} \\

Ours (T=324, 6xBAT) † &\textbf{20.1} & \underline{22.9}&\textbf{19.9}& \underline{20.3}& \textbf{20.6} & \textbf{23.3}& \textbf{22.0}& \textbf{20.0}& \textbf{25.5}& \textbf{27.7}& \textbf{21.4}& \underline{21.8}& \underline{21.1}& \textbf{15.9}& \textbf{15.5} &\textbf{21.2}\\

\hline

\end{tabular}
\label{table:h36p2}
\end{table*}

\subsection{Implementation Details}

\subsubsection{Hyper Parameters and Training}
We conducted an end-to-end training of the proposed model architecture in Fig. \ref{fig:frame}. To balance the number of parameters and the diversity of feature extraction across all model layers, we set the size of the hidden features to $D=32$. The MO-GCN module is designed based on \citet{zou2020high}. In our proposed GOA,  we set the number of orders in the graph as $R=3$, which balances between the computation complexity (graph connections) and the performance. Initially, the BAT structure follows the design in \cite{zheng20213d}. We then incorporated the JWA and BCMA modules into it. We investigate the influence of the number of iterations in
the Transformer encoder (BAT) on the model's performance in the computation complexity section.

We adopted the same training parameters (e.g., Adam optimizer, momentum, initial learning rate, and learning rate decay factor) as used in  \citet{zheng20213d}. However, through trial-and-error, we determined that the optimal learning rate is  0.0003, which is decreased by a factor of 0.95 after each epoch.  Following previous methods,  such as \cite{pavllo20193d, liu2020attention}, and \cite{chen2021anatomy}, we employed the MPJPE loss function, which minimizes the error between the ground truth and predictions. The experiments were conducted using the PyTorch framework on a machine with an Nvidia GeForce RTX 4090 GPU, 64GB of RAM, and an Intel(R) Core(TM) i9-13900k CPU.

\subsubsection{Overfitting Mitigation}
Throughout our experiments, we observed severe overfitting, where the model performs well on the training data but struggles to generalize on the testing data after a few epochs.  To mitigate this, we applied data augmentation through pose flipping, similar to other approaches. We also found that using a smaller batch size of 16 significantly mitigated overfitting.  Additionally, using layer normalization after GOA, as well as before JWA, between JWA and BCMA, and after BCMA, enhances the model's generalization capability. Moreover, incorporating a residual connection between the output of GOA and the MLP output in BAT significantly reduces overfitting. Furthermore, we observed that incorporating the full video frames as input, followed by periodically dropping frames before the first layer of the model, helps alleviate overfitting for large datasets. For a fair comparison, we first use the full video as input and then reduce the frame rate within the model by applying a predefined dropping rate. In the Human3.6m dataset, we use 324 frames as input with a drop rate of 3 frames.  For the MPI-INF-3DHP dataset, we use 81 frames as input with a drop rate of 2 frames. However, in the HumanEva-I dataset, we use 5 frames without applying any frame dropping.

\begin{table}
\renewcommand{\arraystretch}{1.2}
\setlength{\tabcolsep}{3.5pt}
\caption{Comparison with state-of-the-art on MPI-INF-3DHP dataset. † indicates using temporal information. * indicates reconstructing an intermediate 3D pose sequence.}
\centering
\begin{tabular}{l |c  }

\hline
  Method   & MPJPE         \\
\hline
\citet{mehta2017monocular} (3DV'17, T=1)&  117.6\\
\citet{pavlakos2018ordinal} (CVPR'19, T=81) & 84.0\\
\citet{lin2019trajectory} (BMVC'19, T=25) † & 79.8\\
\citet{zheng20213d} (ICCV'21, T=9) †& 77.1\\
\citet{chen2021anatomy} (TCSVT'21, T=81) † & 78.8\\
\citet{shan2022p} (ECCV'22, T=81) † & 32.2\\
\citet{zhao2023poseformerv2} (CVPR'23, T=81) † & 27.8\\
\citet{yu2023gla} (ICCV'23, T=81) †& \underline{27.7}\\

\citet{wang2020motion} (ECCV'20, T=96) †* & 68.1\\
\citet{zhang2022mixste} (CVPR'22, T=27) †*& 54.9\\
\citet{hu2021conditional} (MM'22, T=96) †* & 42.5\\
\citet{gong2023diffpose} (CVPR'23) †*& 29.1\\

Ours  (T=81, 8xBAT) †     & \textbf{24.7} \\

\hline
\end{tabular}
\label{table:mpi}
\end{table}

\begin{table*}
\renewcommand{\arraystretch}{1.2}
\caption{Comparison with state-of-the-art on HumanEva-I DATASET using 2D ground truth (GT) joints as input. † indicates using temporal information. * indicates reconstructing an intermediate 3D pose sequence.}
\centering
\begin{tabular}{l |c c c|c c c|c }

\hline
  \multirow{2}{*}{Method}    & \multicolumn{3}{c|}{Walk.}  & \multicolumn{3}{ c |}{Jog.} & \multirow{2}{*}{ Avg}\\ 
                & S1 & S2  & S3            &  S1 & S2 & S3       &   \\
\hline
\citet{martinez2017simple} (ICCV'17) & 19.7& 17.4& 46.8& 26.9& 18.2& 18.6& 24.6\\
\citet{fang2018learning} (AAAI'18) &19.4 &16.8& 37.4 &30.4& 17.6 &16.3&  23.0\\
\citet{lee2018propagating} (ECCV'18)  & 18.6 & 19.9 & 30.5 & 25.7 & 16.8 & 17.7 &21.5\\
\citet{pavlakos2018ordinal} (CVPR'18) &18.8 &12.7 &29.2& 23.5 &15.4 &14.5&19.0  \\
\citet{cheng20203d} (AAAI'20) † & 10.6&11.8& 19.3& 15.8&11.5&12.2&13.5\\
\citet{zheng20213d} (ICCV'21) †  & 14.4 & 10.2&  46.6&  22.7&  13.4&  13.4 &20.1\\
\citet{yu2023gla} (ICCV'23) †&  \underline{8.7} &\underline{6.8} &\textbf{11.5}& \textbf{10.1} &\underline{8.2}& \underline{9.9}&\textbf{9.2}\\
\citet{cheng2019occlusion} (ICCV'19) †* &11.7 &10.1 &22.8 &18.7& 11.4 &11.0 &14.3\\
\citet{zhang2022mixste} (CVPR'22) †*& 12.7 &10.9 &17.6& 22.6& 15.8 &17.0  &16.1\\
\citet{li2022exploiting} (TMM'22) †*& 9.7& 7.6& \underline{15.8}& \underline{12.3} &9.4 &11.2&11.1\\

Ours  (T=5, 8xBAT) †	  & \textbf{8.7} & \textbf{6.5}  & 17.9  & 13.5 & \textbf{7.8} &   \textbf{8.5}&\underline{10.4}  \\

\hline
\end{tabular}
\label{table:hmeva}
\end{table*}

\subsection{Comparison with the State-of-the-art}   

Existing works on 2D to 3D pose estimation typically distinguish between two types of approaches: temporal approaches and 3D reconstruction methods. Temporal approaches (denoted †), take a 2D sequence as input and output a single 3D pose corresponding to the central frame in the sequence. Examples include works like:  \cite{zheng20213d}, \cite{zhou2021hemlets}, \cite{xu2021graph}, MHFormer\cite{li2022mhformer}, \cite{shan2022p}, GLA-GCN\cite{yu2023gla}(†), PoseFormerV2\cite{zhao2023poseformerv2}. In contrast, 3D reconstruction methods (denoted †*) involve sequence-to-sequence predictions that map each 2D pose in the input sequence to a corresponding 3D pose in the output sequence or reconstruct an intermediate 3D sequence before producing the final single 3D pose output. These methods typically demand higher computational costs and perform better when trained on large datasets compared to smaller ones, such as in the works: \cite{zhang2022mixste}, \cite{hu2021conditional}, \cite{gong2023diffpose}, \cite{zhang2023evopose}, \cite{li2022exploiting},  GLA-GCN\cite{yu2023gla}(†*). These two approaches are compared separately in the literature, such as in GLA-GCN\cite{yu2023gla} and MHFormer\cite{li2022mhformer}.

Our method belongs to the temporal category (†), where the central attention fits to be applied on temporal approaches because it focuses on the central of the sequence. We compare our method with the state-of-the-art temporal methods on Human3.6m, MPI-INF-3DHP and HumanEva-I datasets. However, we also show that our method outperforms some of the 3D reconstruction based methods as well on MPI-INF-3DHP and HumanEva-I datasets.

Table \ref{table:h36p1} shows the comparison with the temporal state-of-the-art methods on Human3.6m dataset using 2D detection (CPN) joints and 2D ground truth joints (GT) as input under $Protocol\#1$ (MPJPE). The results on CPN report results on traditional techniques as well, based on using SIFT for human bounding box extraction and KNN (K-Nearest Neighbors), KRR (Kernel Ridge Regression), LinKRR (Linear approximations for Kernel Ridge Regression) , LinKDE (Linear approximations for Kernel Density Estimation) for prediction. Overall, the performance is on par with state-of-the-art with an average of 43.0, outperforming MHFormer \cite{li2022mhformer} and GLA-GCN\cite{yu2023gla}. It also achieves the highest performance on Eating action with 39.4 MPJPE, and second better performance on Phoning and Waiting actions with 44.5 and 42.0 MPJPE, respectively. On GT, our model's performance is better in average and individual actions except for Discussion, Waiting, Walking, Walking Together, where it shows second second-best performance on Discussion and Waiting. On $Protocol\#2$ in Table \ref{table:h36p2}, we report on par with state-of-the-art performance on CPN  with an average of 34.6, ranked second overall. Additionally, we achieve the best performance on the Phoning and Smoking actions with MPJPEs of 34.7 and 35.5, respectively, and second-best accuracy on the Discussion, Greeting, Waiting, and Walking Together actions. On GT, our model shows the best average accuracy with 21.2 MPJPE and the best performance on individual actions except for Discussion, Greeting, Waiting, and Walking Together, as the second best performance.

Table \ref{table:mpi} shows the comparison of the performance on MPI-INF-3DHP  against state-of-the-art temporal methods and some of 3D reconstruction methods. Our method outperforms recent temporal methods by a significant margin of at least 3. Additionally, when compared to four 3D reconstruction methods (†*), our approach demonstrates superior performance, with a large margin of at least 4.4. While 3D reconstruction methods generally show better performance on large datasets like Human3.6m, temporal methods like ours tend to achieve better results on smaller datasets, such as MPI-INF-3DHP, as shown in Table \ref{table:mpi}.

The comparison of our model’s performance on the HumanEva-I dataset with state-of-the-art methods under $Protocol\#2$, using ground truth (GT) joints, is presented in Table \ref{table:hmeva}. Our method achieves superior results across individual subjects for two actions, except for the Walking action with subject S3 and the Jog action with subject S1. On average, our method ranks second with an overall performance of 10.4.

\subsection{Ablation Study}  
\subsubsection{Effect of Each Component} Table \ref{table:abl} presents the performance of the model in the presence and the absence of the main proposed components, tested on a sequence of 9 frames as input. Using all the components together (MO-GCN, GOA, JWA, and BCMA) generates an accuracy of 35.8 MPJPE (Exp 7), which we consider as the baseline criterion of the accuracy when removing or replacing each of the components from the complete model. 

Omitting the GOA from the model produces a lower accuracy of 37.0 (Exp 1), which demonstrates the positive impact of the attention across the graph orders. Employing  MO-GCN, GOA, and BCMA without the JWA leads to a decrease to 38.6 in performance (Exp 2), highlighting the benefit of utilizing local attention for each joint features over frames instead of relying on global attention only, where all joints are treated as a single entity, which forces the model to learn an averaged representation, losing spatial hierarchy and joint dependencies. Moreover, joint-wise attention enables the model to align motions within the same joint across frames, whereas global attention risks misaligning unrelated joint motions. Furthermore, global attention struggles when certain joints are occluded or noisy, as their features are merged into the other joints' features.

\begin{table}
\renewcommand{\arraystretch}{1.2}
\setlength{\tabcolsep}{5.5pt}
\caption{Ablation study on the effect of each component of the proposed learning architecture (MA: vanilla Multi-head Attention). The results are generated on Human3.6m dataset using 2D ground truth (GT) joints as input with 4xBAT of 9 frames. Exp: Experiment number}
\centering
\begin{tabular}{ l| c | c | c | c | c | c }

\hline
Exp & MO-GCN  & GOA &  JWA  & MA & BCMA &   MPJPE   \\
\hline

1&\checkmark  &   & \checkmark  &  & \checkmark  & 37.0\\
2& \checkmark  & \checkmark  &  &  &   \checkmark &  38.6\\
3&\checkmark &  \checkmark &  \checkmark & \checkmark &  & 38.5 \\

4&\checkmark   & \checkmark   &  & \checkmark  &    &  39.2 \\
5& & &\checkmark   &  &   \checkmark  & 37.8  \\  
6&\checkmark   & \checkmark   &   &  &     & 60.5 \\
7&\checkmark   & \checkmark &\checkmark  &  &   \checkmark & \textbf{35.8}  \\
\hline
\end{tabular}
\label{table:abl}
\end{table}

\begin{table}[h!]
\renewcommand{\arraystretch}{1.1}
\setlength{\tabcolsep}{3pt}
\small
\caption{Effect of the number of training data on the number of frames in the input sequence, and the dropping frame rate inside the model. The numbers in the table correspond to the best accuracy on the three datasets. }
\centering
\begin{tabular}{l | c | c | c  }

\hline
Datasets   & input frames & Drop rate & Training frames\\
\hline
 HumanEva-1 & 5        & 0             & 7419            \\
 MPI-INF-3DHPE & 81   & 3             & 1052800         \\
 Human3.6m & 324      & 4             & 1559752         \\

\hline

\end{tabular}
\label{table:data}
\end{table}

Replacing the BCMA with a vanilla Multi-head Attention (MA) shows a decline in the performance to 38.5 (Exp 3) because of the equal focus on the elements of the sequence, while our BCMA pays higher attention to the center of the sequence. Vanilla MA assigns uniform attention to noisy or redundant frames, while BCMA mitigates this by concentrating on the most informative region. Replacing the entire BAT with the vanilla Transformer encoder (without JWA and BCMA) results in a significant accuracy drop to 39.2 (Exp 4).

To further evaluate the spatial processing components, in Exp 5, we report the MPJPE in the absence of spatial processing, where a simple embedding is used to represent each skeleton sequence instead of employing MO-GCN and GOA. This leads to a noticeable decline in performance to 37.8. Exp 6, which excludes temporal processing, demonstrates the most significant performance drop, with MPJPE increasing to 60.5.

The analysis of the ablation study demonstrates that the proposed modules are crucial to improve the spatial and temporal representation of the skeleton sequence. The spatial skeleton representation is improved with GOA module, and the motion representation is improved via local frame-wise joint attention and global centered attention. The combination of our modules with the existing Multi-order GCN and The Vanilla self-attention showed better performance in terms of joint localization accuracy.

\begin{figure}
\centering
\includegraphics[width=1\linewidth]{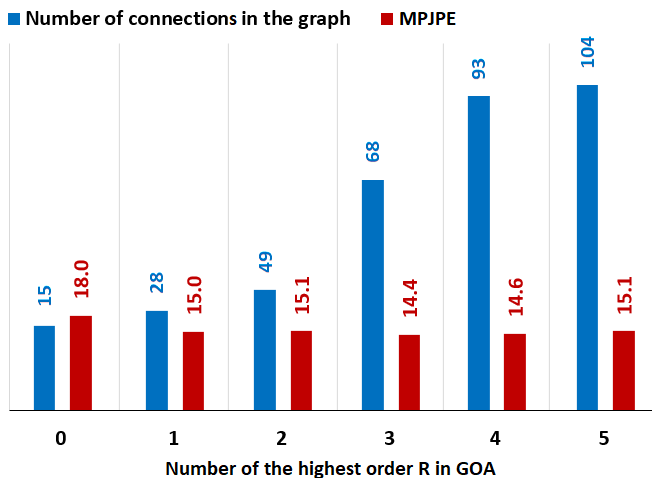}
\caption{Ablation study on the effect of the number of orders R on the performance using 9 frames of 2D ground truth (GT) joints from HumanEva-I dataset on $Protocol\#1$ for 8xBAT.}
\label{fig:ord} 
\end{figure}

\subsubsection{Effect of the Number of Orders} Although we use the same skeleton topology as input, using different orders results in more connections between joints, and hence, better representation. A graph of order $R$ includes the connections of the previous order $R-1$. Higher orders imply more connections between joints. In Fig \ref{fig:ord} we preset the MPJPE of 9 frames on different number of orders applied on HumanEva-I dataset with a skeleton of 15 joints. We notice that the best MPJPE is obtained with $R=3$, and the performance declines with higher or lower order. From  $R=0$ to $R=3$ the number of connections between joints increases from 15 to 68, and the performance increases from  18.0 to 14.4, respectively. However, from  $R=4$ to $R=5$ the number of joint connections increases from 93 to 104, respectively, and the performance decreases. 

This behavior can be attributed to the fact that when the order $R>3$, the number of connections stabilizes because there are no additional sequences of joints supporting the order due to the limited number of joints in the skeleton, which results in the extraction of redundant features, and hence, no more improvement in the performance. In our experiments, we utilize the highest graph order, $R=3$, as it achieves a balance between efficient feature extraction and preventing redundancy.

\subsubsection{Effect of the number of training data}
Table \ref{table:data} shows the number of training data of the three datasets with the number of frames in the input sequence and the dropping rate, that lead to the best accuracy. As we mentioned before, our model takes the full input sequence from the video as input and periodically drops frames based on a specific dropping rate to mitigate overfitting that occurs due to the redundant graph representation of frames, where the poses don't change because of considering the full frame sequence. rate.

For  Human3.6M dataset, which contains the largest training set (1,559,752 frames) and longer video sequences compared to the other benchmarks, our experimental results demonstrate that optimal performance is attained when using an input sequence of 324 frames combined with a dropping rate of 4 frames. This configuration successfully balances the need for sufficient temporal context with the imperative to mitigate redundancy in longer sequences.

The MPI-INF-3DHP dataset presents a different optimization scenario, with its intermediate-scale training set (1,052,800 frames, approximately 33\% smaller than Human3.6M) and moderately shorter video sequences. Our tuning revealed that increasing the input sequence more than 81 frames does not improve the accuracy further, which suggests that this length provides adequate temporal information while avoiding unnecessary computational overhead. The corresponding dropping rate of 3 frames was found to be most effective for this sequence length and dataset.

On the contrary, the HumanEva-I dataset's limited scale (only 7,419 training frames) and short video clips require a much shorter input sequence.  The constrained nature of this dataset means that nearly every frame contains valuable information, leading to our finding that optimal performance is achieved with minimal input sequences of just 5 frames and requires no frame dropping (rate 0).

\begin{table}[h!]
\renewcommand{\arraystretch}{1.1}
\setlength{\tabcolsep}{1pt}
\small
\caption{Computational complexity comparison under $Protocol\#1$ on Human3.6m using 2D (GT) joints with varying iterations of BAT. Results were generated using a single GeForce RTX 4090 GPU.}
\centering
\begin{tabular}{l | c | c | c   }

\hline
     Method    &   frames  & Parameters (M) &   MPJPE    \\
\hline

\citet{pavllo20193d} (CVPR’19)  &  27 &  8.56 &    40.6  \\
\citet{liu2020attention} (CVPR’20)&  27 &  5.69 & 38.9     \\
\citet{li2022mhformer} (CVPR’22)    & 27 &   18.92 & 34.3      \\

\citet{pavllo20193d} (CVPR’19) &  81 &  12.75& 38.7      \\
\citet{liu2020attention} (CVPR’20)& 81 &   8.46 & 36.2     \\
\citet{li2022mhformer} (CVPR’22) &  81 &  $\geq$ 18.92 & 32.7     \\

\citet{pavllo20193d} (CVPR’19)   & 243  &  16.95 & 37.8    \\
\citet{liu2020attention} (CVPR’20)&  243  &   11.25 &  34.7    \\
\citet{li2022mhformer} (CVPR’22) &  351 & $\geq$ 18.9& 30.5    \\

Ours (1xBAT)      & 324 & 2.58&   36.4          \\
Ours (2xBAT)      & 324 & 4.95&    30.4          \\
Ours (4xBAT)      & 324 & 9.70&    27.9          \\
Ours (6xBAT)      & 27 & 14.29&    33.5          \\
Ours (6xBAT)      & 81 &14.32&    32.0        \\
Ours (6xBAT)	     & 324 & 14.45&    27.0           \\

\hline

\hline

\end{tabular}
\label{table:comput}
\end{table}

\begin{figure*}
\centering
\includegraphics[width=0.8\linewidth]{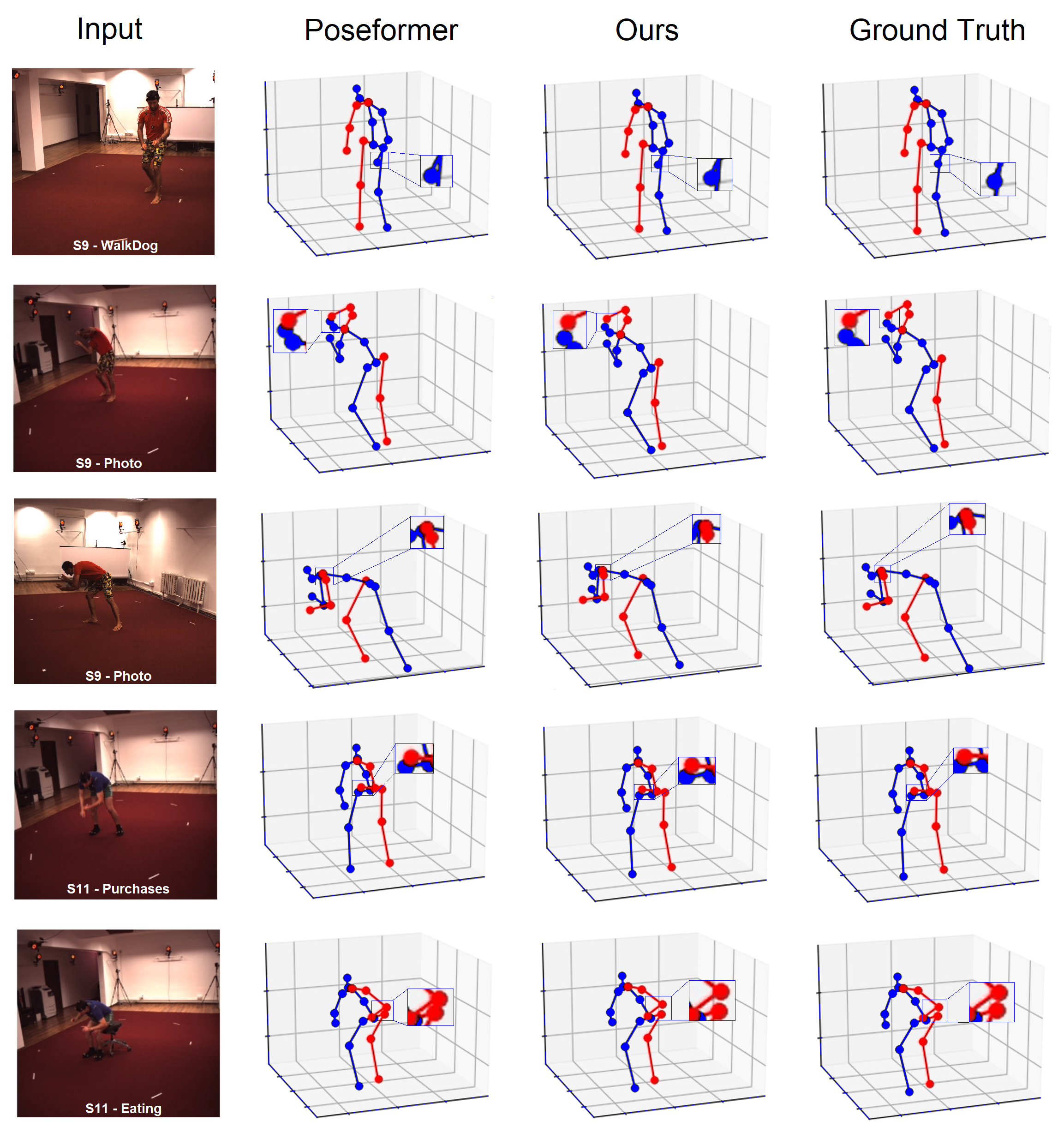}
\caption{Qualitative comparison Poseformer \cite{zheng20213d} and the ground truth for subjects S9 and S11 on 6 actions of Human3.6m dataset using 2D ground truth (GT) joints as input.}
\label{fig:indoor} 
\end{figure*}

\begin{figure*}
\centering
\includegraphics[width=0.9\linewidth]{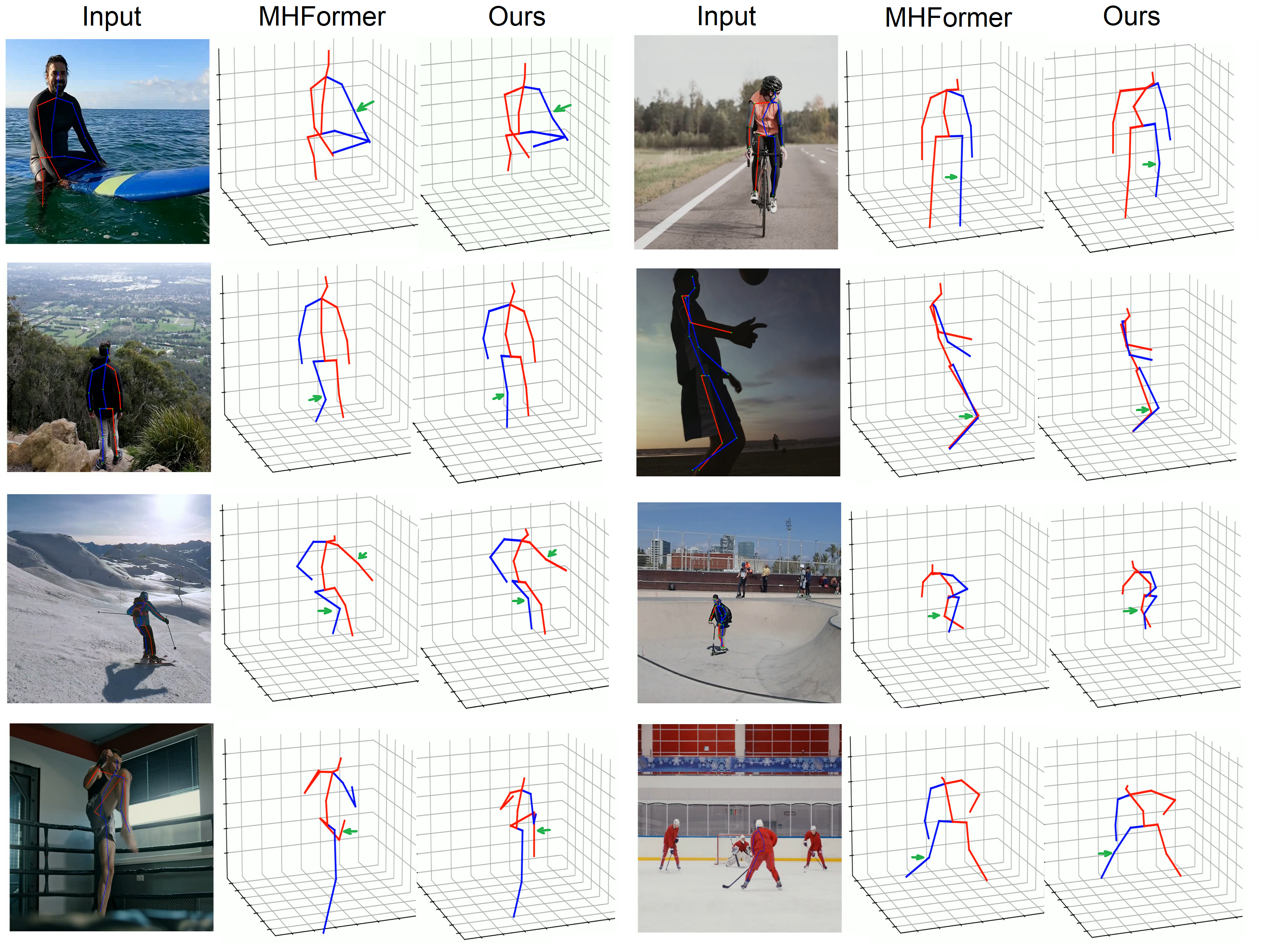}
\caption{Qualitative comparison with MHFormer \cite{li2022mhformer} in the wild. Results obtained from the model trained on the Human3.6m indoor dataset.}
\label{fig:wild} 
\end{figure*}

\begin{figure}
\centering
\includegraphics[width=1\linewidth]{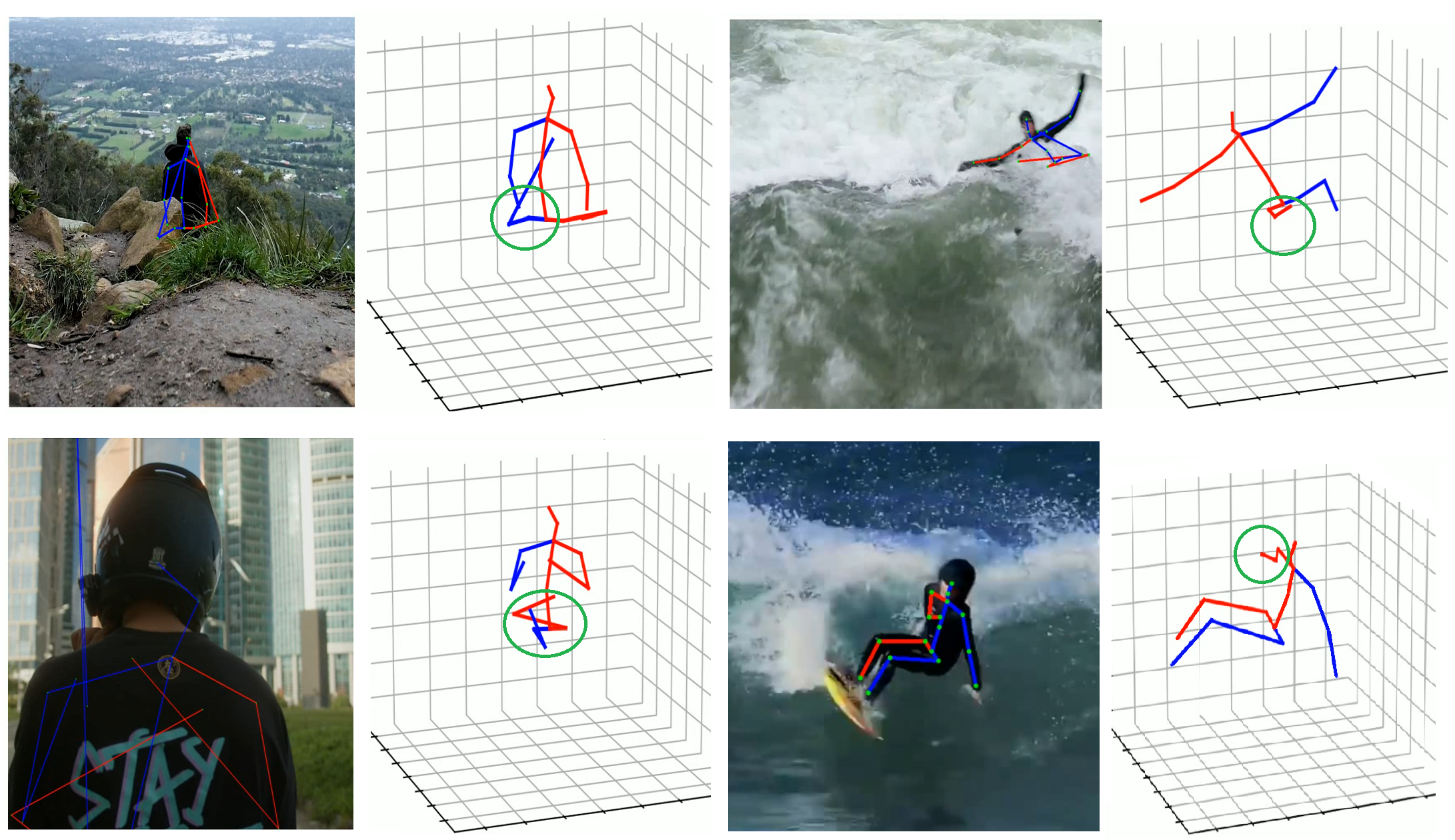}
\caption{Failure cases with different conditions  in the wild generated by our model.}
\label{fig:cases} 
\end{figure}
\begin{figure}
\centering
\includegraphics[width=1\linewidth]{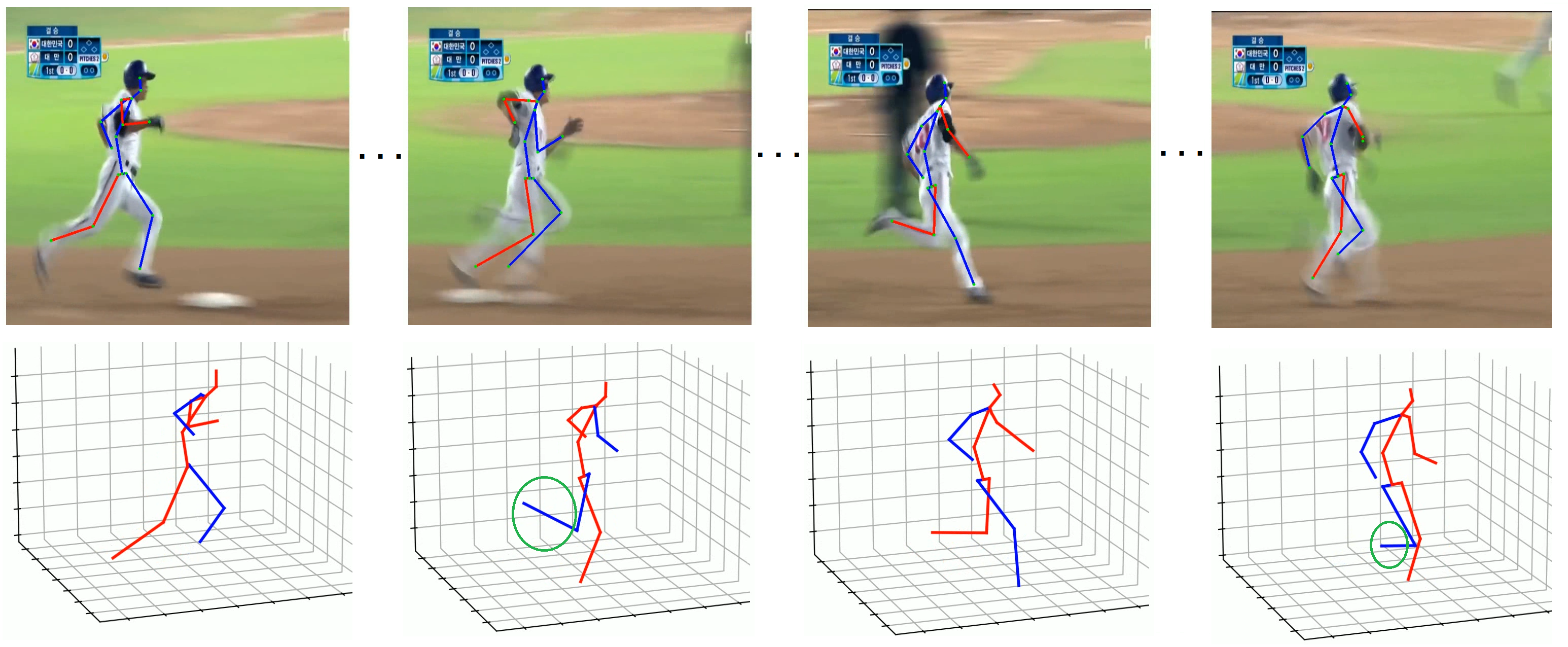}
\caption{Failure pose estimation frames in case of fast movement with blurring.}
\label{fig:speed} 
\end{figure}
\subsubsection{Computation Complexity}
To evaluate the computation cost compared with other methods,  we  present the MPJPE results along with the number of frames and model parameters in Table \ref{table:comput}. We observe that the number of iterations in the temporal transformer encoder (BAT) primarily affects the number of parameters. However, the number of frames does not impact computational complexity.  For a fixed number of BAT iterations, the performance improves as the number of frames increases. Specifically, using 6 iterations of BAT results in higher performance compared to other methods for 27, 81, and 324 frames with 14.29M, 14.32M, and 14.45M parameters, respectively, which is higher than those reported by \citet{pavllo20193d} and \citet{liu2020attention}. However, our model with just 2xBAT with 324 frames shows higher performance than the three methods in terms of computation and prediction accuracy. Furthermore, with only 1xBAT iteration and 2.58M parameters, our model still outperforms \citet{pavllo20193d}. As an advantage our approach, unlike the compared methods, the number of parameters remains stable regardless of the number of frames.

\subsubsection{Qualitative Results}
Fig. \ref{fig:indoor} shows qualitative comparison between the pose prediction of our model and both the ground truth and Posformer \cite{zheng20213d}. The compared joints are highlighted with higher resolution. For a fair comparison, our predictions are obtained on 81 frames as Posformer. To test the model in different situations, the poses are selected from different actions of the testing subjects S9 (Directions, Photo, WalkingDog) and S11 (Purchases, Eating, SittingDown) of the Human3.6m dataset. In general, our model performs better, and the joint predictions are closer to the ground truth than Posformer, especially in challenging poses such as occlusion in S11:Purchaces, S11:Photo, and S9:Directions. The Poseformer model employs two transformers, one for spatial feature extraction and another for temporal modeling, while we use a spatial GCN instead to model each skeleton, which shows that a combination of GCN for skeleton modeling and a Transformer for handling temporal dependencies offers a more efficient solution. 

Fig. \ref{fig:wild} shows the model's performance in out of distribution scenarios, where it is trained on Human3.6m dataset in indoor scenarios and applied to arbitrary videos with different conditions such as lighting, occlusion, and distance from the camera, compared to MHFormer \cite{li2022mhformer} using their provided code. Overall, our model performs as well as it does indoors and surpasses MHFormer in some situations. Specifically, in the case of the picture of the second row (left), where distinguishing between the two legs is challenging due to the darkness, and in the case of the picture of the last row (left), where the joints of the left hand and face are occluded by the knee. 

Additionally, in Fig. \ref{fig:cases} we highlight some failure cases in very challenging conditions, such as extreme proximity to the camera and heavy occlusion, and in Fig. \ref{fig:speed}, we show the performance of the model in the case of very fast body motion of a running person, where the body parts are blurred due to rapid movement. As a result, the model fails to accurately predict the pose in some frames.

\section{Discussion and Conclusion}  

The proposed method in this paper addresses limitations in existing temporal methods that predict
a single output corresponding to the center of a sequence. These methods typically use a basic
Graph Convolutional Network (GCN) or a Transformer for spatial modeling, and another Transformer
with simple attention mechanisms for temporal modeling. However, these approaches often rely on simplistic attention mechanisms that do not effectively capture the complex relationships between joints.

The proposed method introduces a new Spatial GCN that integrates multi-order feature extraction with a Graph Order Attention (GOA) module. The GOA module dynamically assigns different levels of attention to joints based on their order, which enhances the efficacy of spatial skeleton representation. This allows the model to capture complex relationships between joints and better represent the underlying 3D pose. The Spatial GCN is designed to learn robust features by aggregating information from neighboring joints in the skeleton hierarchy. The GOA module takes into account the order of the joints, allowing it to selectively focus on certain joints or ignore others based on their relative position in the sequence.

The method also improves temporal modeling by introducing a new Body-Aware Transformer (BAT) that
captures dependencies in the feature sequence. The BAT employs Joint Weighted Attention (JWA) to
track the evolution of each joint over time, followed by an enhanced self-attention mechanism that
is sensitive to the central elements of the sequence. The JWA module allows the model to selectively focus on joints that are more relevant to the current 3D pose estimation task. The enhanced self-attention mechanism takes into account the
temporal dependencies between joints and the relative position of each joint in the sequence,
allowing it to accurately predict the 3D pose at the center of the sequence.

As shown in the ablation study, replacing the BAT with a traditional Transformer resulted in a significant decline in performance, indicating that the BAT is crucial for capturing temporal dependencies in the feature sequence. Experiments were conducted on three benchmark datasets: Human3.6m, MPI-INF-3DHP, and HumanEva-I. The results show that the proposed method outperforms existing temporal-based methods that use basic GCNs or Transformers for spatial modeling, as well as those employing standard Transformers
with simple attention mechanisms for temporal modeling.

While the proposed method achieves state-of-the-art performance on some datasets, it has limitations. The BCMA module's focus on the sequence center makes it unsuitable for sequence-to-sequence prediction. Additionally, increasing the number of Transformer iterations or using higher orders in the GOA module raises computational cost. 

As future work, we aim to develop a novel attention mechanism tailored to sequence-to-sequence modeling and explore optimization strategies to reduce the computational cost of both the spatial GCN and temporal Transformer.

\section*{Acknowledgement}

Aouaidjia Kamel and Chongsheng Zhang are supported
in part by the National Natural Science Foundation of
China (No.62250410371) and the Henan Provincial Key R\&D
Project (No.232102211021).










\bibliographystyle{cas-model2-names}

\bibliography{cas-refs}

\end{document}